# Multi-agent robotic systems and exploration algorithms: Applications for data collection in construction sites

### S. A. Prieto [a], N. Giakoumidis [b,c] and B. García de Soto [a]

[a] S.M.A.R.T. Construction Research Group, Division of Engineering, New York University Abu Dhabi (NYUAD), Experimental Research Building, Saadiyat Island, P.O. Box 129188, Abu Dhabi, United Arab Emirates.
E-mail: samuel.prieto@nyu.edu, garcia.de.soto@nyu.edu
[b] KINESIS Lab, Core Technology Platforms, New York University Abu Dhabi, United Arab Emirates.
[c] Intelligent Systems Lab, Cultural Technology and Communication, University of the Aegean, Greece.
E-mail: giakoumidis@nyu.edu

## Abstract

The construction industry has been notoriously slow to adopt new technology and embrace automation. This has resulted in lower efficiency and productivity compared to other industries where automation has been widely adopted. However, recent advancements in robotics and artificial intelligence offer a potential solution to this problem. In this study, a methodology is proposed to integrate multi-robotic systems in construction projects with the aim of increasing efficiency and productivity. The proposed approach involves the use of multiple robot and human agents working collaboratively to complete a construction task. The methodology was tested through a case study that involved 3D digitization of a small, occluded space using two robots and one human agent. The results show that integrating multi-agent robotic systems in construction can effectively overcome challenges and complete tasks efficiently. The implications of this study suggest that multi-agent robotic systems could revolutionize the industry.

## 1 Introduction

Construction is one sector that has not fully utilized automated processes during onsite operations [1]. Unlike other sectors of the industry that integrated automated processes for their operations, the lack of automation in construction impacts its efficiency and productivity. Several studies show that automation in construction would significantly improve the efficiency and productivity of the overall life-cycle of the construction process [2]. One of the reasons behind the slow adoption of automated techniques can be attributed to the challenging nature of construction projects. In general, construction sites are constantly evolving and dynamically changing states, making it extremely difficult for an autonomous platform (i.e., a robot) to perform in such uncontrolled environments [3].

Another challenge for the construction field adopting automated techniques is the vast amount of different individual tasks required in a single construction project [1], [3]. Robots with a single-purpose task need to be designed to fulfill that specific task and cannot be used for other purposes. Compared to single-purpose robots, general-purpose robots can perform general operations that involve a set of individual sub-tasks. General-purpose systems are more likely to be adopted in the construction field for their wider range of applications and possibilities [4], such as general-purpose applications (e.g., autonomous data acquisition [5], [6], progress monitoring [7] or quality assessment [8] of the construction process). One of the benefits of general-purpose systems is that the different tasks involved in the process can be allocated and performed in parallel by multiple robotic agents, creating opportunities to increase the overall efficiency and robustness of the process. The collaboration of multiple agents presents some challenges that need to be addressed, such as task allocation and communication between the different agents. In order to solve the task allocation problem, a set of different roles with a clear set of instructions need to be developed to provide the necessary structure to the system.

To take advantage of the potential benefits associated with the use of and collaboration among different robotic systems in construction projects, we propose a methodology to facilitate the integration of multi-agent robotic systems in the construction field. The methodology is developed for a general-purpose application, presenting a particular case study of a multi-agent robotic system with the goal of autonomous data collection in a construction site. The data acquisition process is based on an autonomous exploration approach, taking advantage of the two different roles assigned to the different robots with different payloads and capabilities.

The rest of the paper is structured as follows: Section 2 goes over the current state-of-the-art multi-agent robotic systems, as well as the different approaches used for navigation with an emphasis on autonomous exploration. Section 3 presents the proposed methodology and explains its key elements. Section 4 shows a case study to illustrate the implementation of the proposed methodology. Section 5 discusses the findings and lessons learned from the case study and points out some of the limitations of this study. Finally, Section 6 concludes the paper and provides an outlook for future work.

## 2   Previous work

### 2.1   Multi-agent robotic systems

Research has proven that developing a multi-agent robotic system (MARS) is more cost-effective than developing a single costly robotic platform with all the capabilities [9]. When it comes to defining the taxonomy and architecture that different MARS can adopt, there are systems where the workload and task assignment are equally distributed between the different agents inside the MARS, and systems where there is a hierarchy between the different agents and one of them is acting in command [10]. With systems where the number of agents is minimal, the latter approach is more efficient and robust. Within the context of MARS, there are at least two agents, each with different roles. In general, one of them is in charge of assigning tasks, and the other(s) would execute the given commands. This architecture has been widely used in the medical field, where surgical robots and human surgeons act as one system. In this example, the human surgeon with high-level decision-making skills commands the robot through an interface device that executes the commands with high precision. Generally, many simple and repetitive tasks can be addressed efficiently with MARS because of the capability to work in parallel with many tasks.

Construction sites are characterized by intense interaction among multiple agents. For example, in most construction processes, different agents have different tasks in the same environment to complete the process, making these good candidates for a MARS. A few approaches where MARS has been used for construction applications include mapping the environment, block placing, or 3D concrete printing coordination [11]–[13]. A MARS consisting of four robots was used in a tunnel disaster investigation field experiment at the National Institute for Land and Infrastructure Management in Tsukuba (Japan). The experiment showed the viability of the multi-robot approach for inspecting disaster areas [14]. Prieto et al. [8] proposed a concept for using a MARS to collect data automatically that could be used for quality reports. The system consisted of two robots with data acquisition sensors with different levels of quality. When additional information would be needed, the main robot (with low quality but high range) would command the second robot (with high quality but low range) to approach the area of interest to collect more detailed data. Xiao et al. [15] gave some examples of the MARSs applied to construction and summarized some studies addressing the importance of communication capabilities in MARSs [16]–[18].

### 2.2   Different navigation approaches

For a robot to autonomously navigate an environment, three things are needed [19]: (1) a map of the environment and (2) the ability of the robot to be able to localize itself on the map, and (3) the ability to perceive the environment. In order to achieve that, extensive research has already been done in the field of robotics [3], [20]–[22]. The different ways to solve this issue can be categorized into three main approaches, varying in different degrees of autonomy.

The first one, with less autonomy, assumes that the map of the environment is already known by an external source, and the system only has to deal with the localization. The map can come from different sources, although the most common is an existing site floor map. In the construction field, Building Information Modelling is generally well-adopted and used in most construction projects, and most of the information regarding the site is already present in the (Building Information Model) BIM. Yu et al. [23] present a 2D exploration approach based on chaotic exploration and thinning topological mapping. Their method assumes that a 2D map of the scene is already known. Their study focuses on the optimal traversing and navigation through the environment to provide a fast, reliable exploration and the optimal trajectory to return after the task has been completed. In the study by Giusti et al. [24], a robotic platform was developed during the COVID-19 outburst to autonomously disinfect buildings. The location and information provided to the robot about the components to be disinfected were extracted from the BIM. The BIM was also used to describe the environment in which the robot operates without requiring any mapping procedure. They propose an integration between BIM and the Robot Operating System (ROS) through the ROSBIM module, although not much information is provided regarding this module. Another approach using the ROSBIM integration is the one of Kim et al. [25]. They propose an autonomous robot for indoor wall painting, obtaining all the scenario information from the BIM. From the IFC (Industry Foundation Classes) file, the geometry is extracted and converted into ROS-compatible simulation files to simulate a case study where the robot autonomously deals with the task allocation needed to paint the indicated walls. Their proposed method is only tested in simulation and can only be applied to planar surfaces. Karimi et al. [26] proposed a new standard, Building Information Robotic System (BIRS), meant to help the transfer of semantic information from the BIM and Geographic Information System (GIS) to the robotic platform. In their study, they develop a semantic web ontology to integrate the robot navigation and data collection to convey the meanings from BIM-GIS to the robot. Despite claiming their robot to autonomously navigate from one point of the building to another, their experimentation does not provide enough proof to support that claim. Another approach using the BIM information, in this case for progress monitoring and quality assessment purposes, is the one of Ilyas et al. [27]. This information is used to generate a map to be used by the robot for navigation, also providing information regarding the position of the different elements to be inspected. Since they solely rely on the BIM information for the navigation, the robot might have localization problems if the environment has been heavily furnished or the as-built is not exactly as the as-plan. Yin et al. [28] used the BIM to obtain a 2D map that was later used with an Iterative Closest Point (ICP) approach in conjunction with the sensor data to provide robot localization. These approaches have the advantage that the robotic system already knows the map, and the task allocation can be done beforehand, ensuring that the whole process is performed more efficiently and faster. However, prior knowledge of the scene is needed, which might not always be available. On top of that, the as-is state of the site might differ from the as-planned design present in the BIM. This means that the environment the robotic system will end up navigating might be different from the map it will be using, which could lead to localization errors if the robotic platform is not providing real-time feedback to the map obtained from the BIM.

The second category uses the robotic system to generate a map of the physical environment. This is done utilizing Simultaneous Localization and Mapping (SLAM) approaches. The robotic system needs to be driven (i.e., remote-controlled) around the area of interest to generate a map providing full site coverage. This process can be time-consuming and is subjective to the skills of the user operating the platform in combination with the specifications of the depth sensors. Xue et al. [29] proposed an improvement on the LeGO-LOAM based SLAM approach to generate a 3D representation of an underground coal mine. They managed to improve the accuracy of the generated map by using ICP to fuse the results from the LeGO-LOAM module and a SegMatch algorithm. Another SLAM-based study is the one proposed by Chen et al. [30]. They proposed a robotic platform for recycling construction and demolition (C&D) waste and used a SLAM-generated 3D map combined with the data from a 3D LiDAR and visual odometry from an RGB-D camera. Once the environment is explored, the robot autonomously patrols the known area to identify waste that can be picked up with a robot arm and collected in a garbage container. With the aim of 3D digitizing an outdoor construction site, Kim et al. [31] built an initial 2D map of the site from a voxelized representation of the environment. The voxelized representation is obtained through Structure from Motion (SfM) using a UAV manually collecting aerial images. With the 2D map generated, they compute an optimal path planning for an Unmanned Ground Robot (UGV) to stop in the predefined locations to collect 3D data. Asadi et al. [22] presented a monocular SLAM approach to creating a 2D spatial map of the scene. The generated map also presents semantic information about the detected obstacles through a context awareness module based on an ENet Deep Neural Network. Overall, using a SLAM-based approach removes the need to have prior knowledge of the scene; however, it still maintains a manual aspect that might not be desirable in all applications.

The third category also uses a SLAM approach to generate the map of the unknown environment. The difference is that the robot is not manually commanded but autonomously driven by an exploration approach. Exploration algorithms can analyze the space seen by the robot, obtaining the traversable space within the known space and computing the most optimal position for the robot to travel to keep unveiling the unknown space. Most exploration algorithms focus on processing the 'frontiers' of the known space [32]. A distinction can be made based on the type of data (either 2D or 3D) that the algorithm is processing. One of the early studies on exploration algorithms is the one by Nagatani et al. [33]. They presented a Generalized Voronoi Graph (GVG) for mapping and localization. The proposed GVG approach removes the need to update encoder values since it is based on line-of-sight information from the sensors. Their approach mainly focuses on the localization problem, and they do not study how to make the mapping process more efficient and reliable. Marie et al. [34] used data from a 2D monocular catadioptric camera to perform image processing to get the frontiers of the known space. The visual free space is extracted from the skeleton of the local space surrounding the robot, which is used to progressively explore the space and build a map. The path is based on selecting a single local skeleton branch with the closest angle to the current robot heading. The unvisited branches are not considered for further exploration, resulting in an incomplete explored map. Jang et al. [35] proposed an approach of multiple UAVs working together to explore and gather as much information as possible from an unknown scenario. Their exploration method is based on Gaussian Process (GP) regression, which can provide probabilistic inferences over the entire space. The result of the process is an uncertainty map of the occupied space within the environment, which could be used for navigation purposes in a non-cluttered environment. On the other hand, there are 3D-based exploration algorithms like the one proposed by Eldemiry et al. [20]. They developed an autonomous exploration approach for high-quality mapping using feature-based RGB-D SLAM. Their approach relies highly on the quality of the textures used as input, which would determine how many features the algorithm uses to close the loop. As can be seen from the literature review, exploration algorithms are well-researched. However, their application to the construction field has been very limited.

The first category of approaches presents a significant amount of uncertainty when operating in unknown environments, such as dealing with dynamic or unknown obstacles or the possibility of the construction site not looking like the floorplans. Pure SLAM approaches, as in the second category, solve that uncertainty by providing real-time data from the site, but they depend significantly on teleoperation. Exploration approaches add the autonomous aspect missing in the second category, making them the most suitable technique to collect data autonomously from dynamic environments such as construction sites. However, most of the reviewed approaches lack interoperability and scalability, making them hard to implement in large multi-agent systems. Most current exploration algorithms have been developed and tested in controlled and simulated environments, and their robustness to real-world conditions has not been proved. In addition, exploration approaches can benefit from incorporating prior knowledge about their environment, such as maps or semantic information. The approach presented in this paper aims to fill in some of the major gaps in the current research by introducing a robust and scalable methodology for multi-agent systems that use exploration techniques and previous knowledge present in the BIM. By using the exploration technique, the system is robust and reactive to changes in the construction site. By adding known information from the BIM, the exploration stopping criterion can adapt to changing environments, allowing the system to have prior insight for determining the true extent of an unknown environment.

## 3    Methodology

To promote and maximize the use of autonomous robotic systems applied to construction, a methodology that integrates multiple agents and reduces the level of human interaction for all the related elements (i.e., from navigation to task-specific goals) is proposed. The methodology (Figure 1) is broken down into the different agents involved, each with different functions and capabilities, and uses exploration algorithms to navigate the environment.

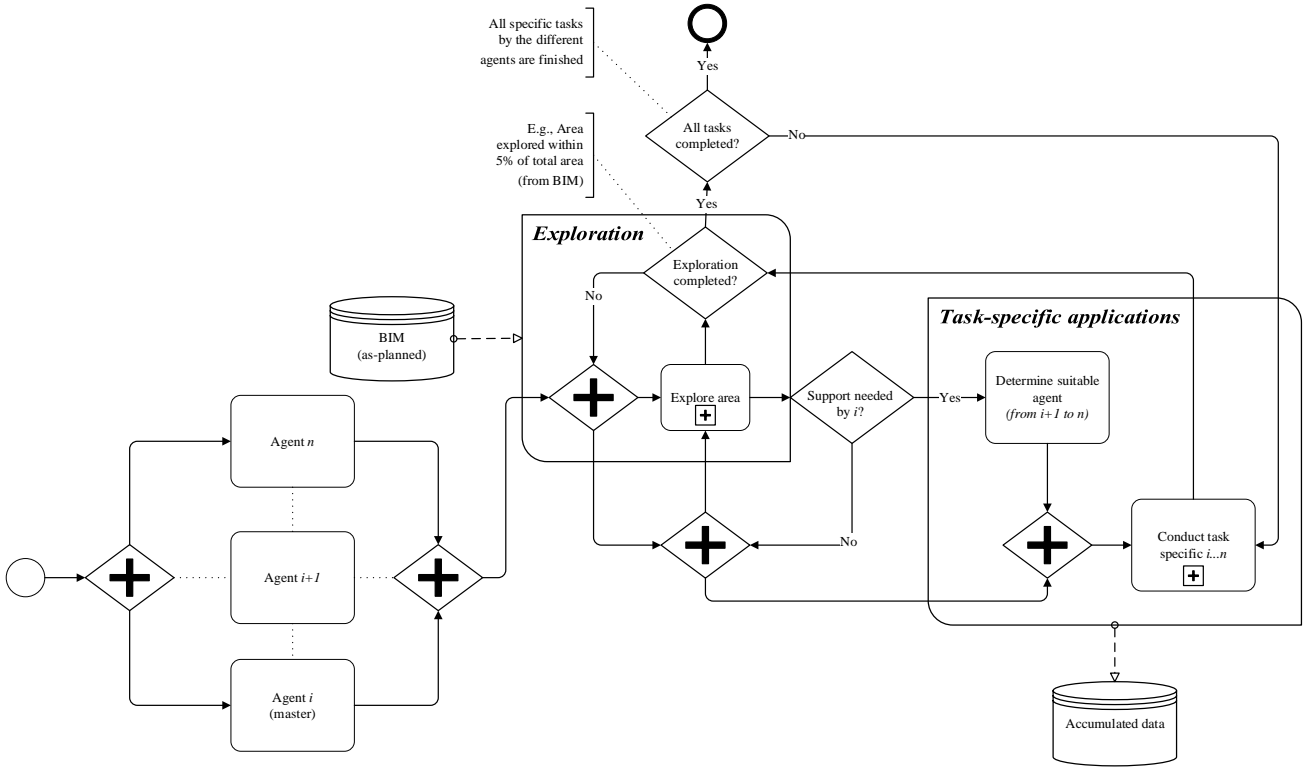

Figure 1. Flowchart of the proposed methodology for the integration of multi-agent systems using exploration.

## 3.1   System Components

### 3.1.1   Control Monitor Station (CMS)

The Control Monitor Station (CMS) is the primary interface between the human agent (HA) or administrator and the Multi Robotic Agent System (MRAS). The administrator can monitor and control the system's operation and each robotic agent through the CMS.

The control of the system can be broken down into high-level and low-level commands. The high-level commands include start-stop operation, new tasks list, overriding a procedure, etc. The low-level commands include manual teleoperation of the physical robot and manual operation of the payloads, usually needed in case of a failure or a special mission.

The monitor of the system is split into two categories; the current status of the system's operation, such as alerts, tasks progress, operation information, etc., and the data output of the system collected from the payloads (e.g., sensors, cameras, 3D scanners, etc.) of the robotic agents.

### 3.1.2   Robotic Agents (RA)

The Robotic Agents (RA) are comprised of four building blocks: the mobile robot, computation unit, communication module, and payloads.

The mobile robot can have any form of locomotion (i.e., wheeled, quadruped, aerial robot, tracked) necessary to reach the physical space for completing an allocated task. The computation unit is responsible for essential operations of the RA, such as navigation, data acquisition, onboard data processing, etc. Lastly, the communication modules interconnect all MRAS nodes (CMS, RA, HA). A multi-node system such as the MRAS operating in a complex and dynamic environment such as a construction site requires a sophisticated communication method.

The communication modules form a wireless communication system based on a Mobile Ad-Hoc Network (MANET). The communication modules create a long-range, self-formed, self-healed, high bandwidth network capable of keeping the robots that are streaming a large amount of data, such as 3D maps, photos, videos, measurements, etc., interconnected. The RA can carry additional communication modules as payloads, which can be placed in critical positions to extend the communication range or overcome an impenetrable obstacle.

### 3.1.3  Human Agents (HA)

The ultimate goal of this project is a complete autonomous mission execution of the MRAS. The MRAS relies on the RA to complete subtasks to achieve the overall mission. There are cases that the needed task exceeds the capabilities of the RA; for example, a big load of raw materials is completely blocking the way of the RA, and there is no path to reach their goal location. In these cases, human intervention is needed to complete a subtask, so the Multi Robotic Agent System continues the mission. The HA can be considered part of the system, but it is not necessary for it to be aware of the overall mission of the Multi Robotic Agent System or the subtask of the RAs. The HA receives a subtask request through a Human Interface Device (HID).

## 3.2  Explore area

To boost productivity and make the system more efficient, all the agents will explore the area simultaneously as they move through the space. This will allow for fast development of the navigation map. While exploration is being done, the agents will perform the assigned tasks. The process for exploring the area is explained next.

### 3.2.1  Floor plan extraction

Most exploration algorithms require some input to define the stopping criteria. For the robot to know when all the available space has been explored, the proposed approach uses the already existing information within the BIM. One of the most common file formats to represent the BIM is the Industry Foundation Classes (IFC) protocol. The IFC file contains the BIM information in a tree-based structure that is easy to parse and retrieve. However, not all IFC files contain the same information. This is very dependent on the software used to create the BIMs and to export the IFC file. In order to provide a robust solution that works no matter what type of information is present in the IFC file, the approach presented here uses the meshed geometry information. The entire mesh of the building can be retrieved using the 'ifcopenshell' library [36]. After that, a horizontal slice is cut from the mesh and projected into a 2D plane, obtaining an initial 2D map representing the entire floor of the building that can be used as ground truth for the exploration algorithm (Figure 2). This can be done for each building floor used in the application. More details regarding this approach can be found in [37].

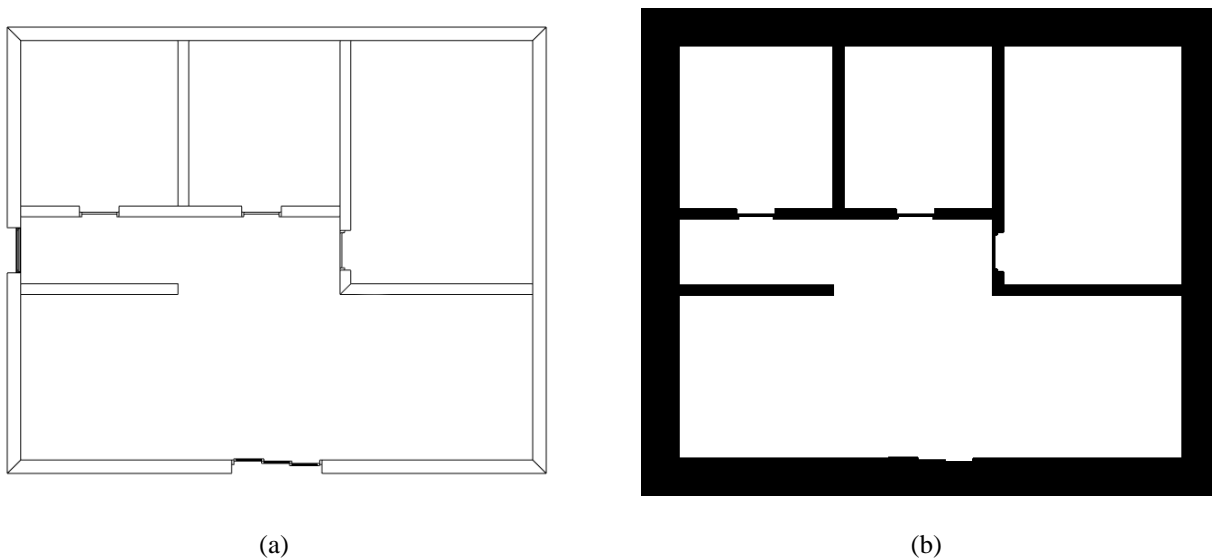

(a)                                                          (b)

Figure 2. (a) Generated 2D map from the IFC file and (b) binary representation of the available space for exploration.

### 3.2.2  Autonomous exploration and stopping criteria

The information extracted from the IFC file can be used to compute the stopping criterion for the exploration algorithm. By providing the exploration algorithm with prior information from the site, the approach gains robustness and stability regarding how much of the available space is still unknown for the system. In most approaches, the exploration process continues until the system concludes that all the space has been explored. The way the system evaluates this is based on the availability of navigable space. This means that if the robotic agent has nowhere to go (i.e., there are no available frontiers between known and unknown space), the system will assume all the space has been explored. However, this approach lacks robustness and could leave unexplored areas that were not initially accessible by the robotic agent. In the presented methodology, prior information from the BIM is considered, and the stopping criterion for the exploration algorithm to stop is mainly based on the percentage of the area explored with respect to the total area represented in the as-is 2D map (i.e., ground-truth). The stopping criterion can be defined based on the findings from the comparison, as indicated by the different cases shown below.

<u>Case 1: The space has not been fully explored yet</u>

In this case, the stopping criterion based on the area percentage has not been fulfilled, and the robotic system needs to keep exploring. As stated before, the main aspect to be considered while evaluating the stopping criterion ($\varphi$) is the amount of area explored with respect to the total available area from the previously extracted as-planned floor map. If the ratio between the two areas (Eq 1.) is higher than a set threshold, the exploration algorithm will finish. The total area explored by the algorithm ($\sum P_{exp}$) is computed by the sum of all the pixels labeled as "free" in the generated map. In the same way, the total area to be explored in the ground truth ($\sum P_{GT}$) is computed by the sum of all the pixels labeled as "free".

$$\varphi = \left(\frac{\sum P_{exp}}{\sum P_{GT}}\right) \times 100\% \qquad (1)$$

If the agents do not encounter any obstacle or impediment while exploring the space, the system will continue to explore until the stopping criterion ($\varphi$) is met. In order to keep exploring, the agents compute the frontiers between the known and the unknown space, computing the centroids of said unexplored frontiers. Those centroids are evaluated, and the robot chooses the most optimal one as its next goal. In Figure 3, an intermediate stage of the exploration process can be seen. The frontiers computed between the known and unknown spaces are represented in blue. The exploration algorithm sends the next goal to the robot, indicated by a red arrow. The navigation stack uses the map generated by the SLAM algorithm to compute a set of costmaps (global and local) represented in the figure with different shades of blue.

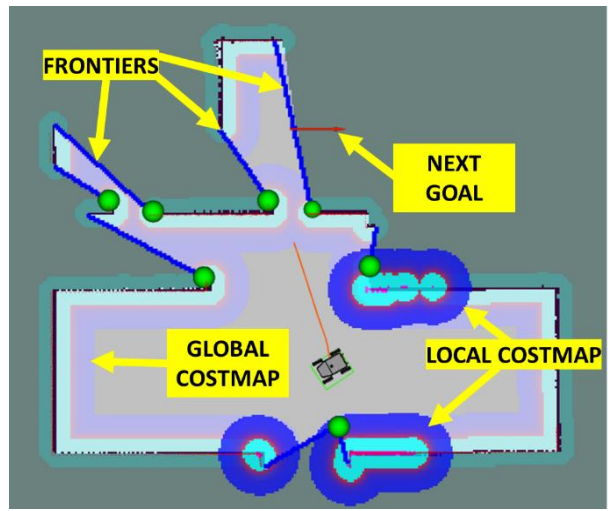

Figure 3. Example of a situation where the scenario has not been fully explored. The unexplored frontiers appear in blue. Annotations of the different elements are highlighted in yellow.

As long as the stopping criterion ($\varphi$) is not fulfilled, the robotic agent will continue to move autonomously towards the available frontiers to keep exploring the space. In Figure 4, four different stages of an exploration process can be seen. As the robot moves towards unexplored frontiers, the percentage of explored area with respect to the ground-truth area keeps growing. In the last step, even though there are still available frontiers for the robot to evaluate, the system stops as the stopping criterion has been fulfilled ($\varphi > 99\%$).

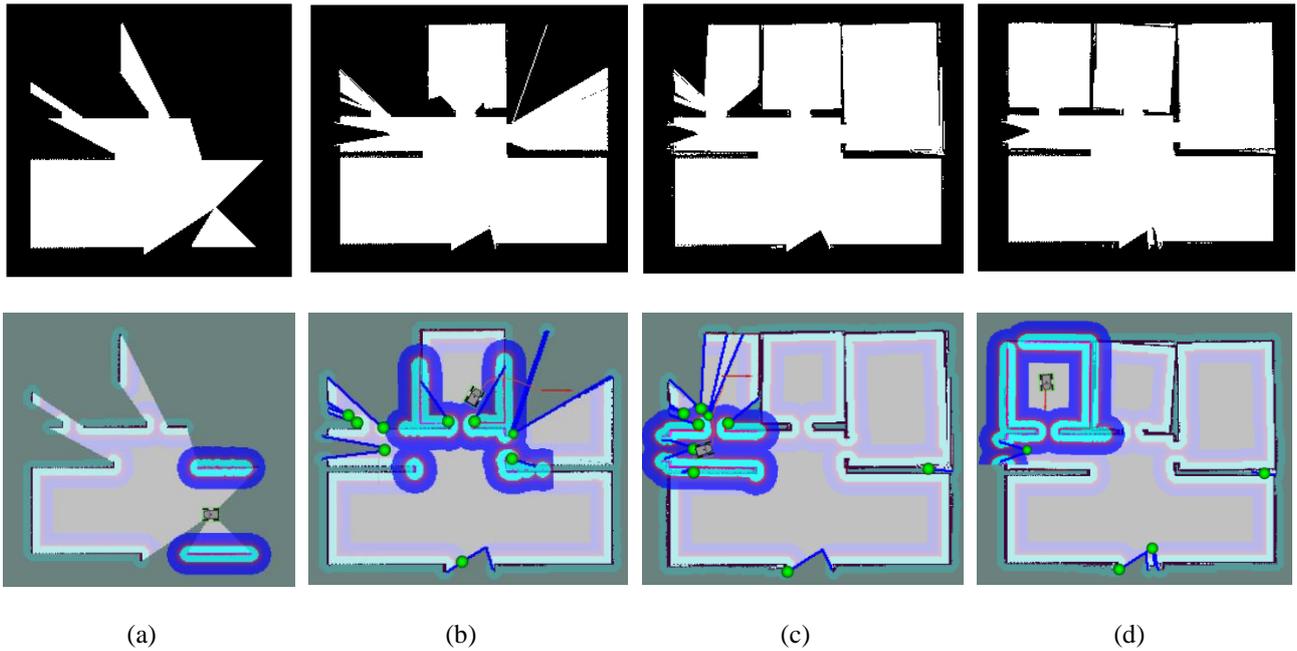

(a)                          (b)                          (c)                          (d)

Figure 4. Binary map of the explored space (upper row) and robot view (lower row) at different stages of the exploration process. (a) stopping criterion of 43.15%, (b) stopping criterion of 75.87%, (c) stopping criterion of 97.15%, (d) stopping criterion of 99.24%.

Case 2: The missing space cannot be accessed, or it is blocked

If the exploration map is still missing information with respect to the as-is floor, then it means that access to the unknown space is blocked. The space can be blocked by obstacles (Figure 5) preventing the robot from navigating toward the unknown environment (i.e., some areas in a given space are enclosed by walls and a closed door, or there might be some elements on the path of the robot that does not allow it to proceed). In this case, the stopping criterion has not been fulfilled, and the robotic system needs to find a way to continue the exploration. In such a situation, the robotic agent needs to send a warning indicating that there is additional space, but it is not accessible. This will trigger a message letting the other agent(s) know that help is needed in order to progress with the autonomous exploration. By accessing the semantic information of the as-is floor map, the location and type of connection between the spaces (i.e., door type, hinge location, pull or push) can be extracted, and a suitable agent can check if the space is indeed blocked by a closed opening or an obstacle. In such a case, the chosen agent should be equipped with a manipulator (e.g., a robot arm), sensing capabilities (i.e., a high-resolution RGB-D camera) and enough computing power to run Artificial Intelligence algorithms to confirm the nature of the object blocking the access to the space (e.g., a door, debris, stored materials). If that capability is not available, it would be expected that remote operation should allow the human agent to intervene. After determining the nature of the blockage, the human agent can provide access to the other agent(s).

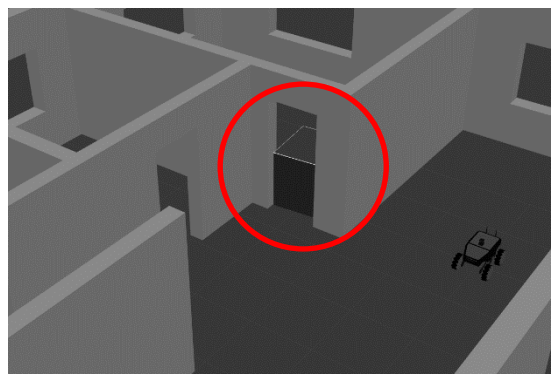

Figure 5. An obstacle in one of the doors (identified by the red circle in the figure) blocks access to one of the available spaces.

An example of this situation is displayed in Figure 6. The robotic agent performs exploration as expected, like in the previous case. However, at some point (Figure 6d), there are no more available frontiers for the robot to continue exploring. A usual exploration approach would conclude here and consider the space completed (i.e., fully explored). However, since the system has prior information from the IFC, the area explored is only 79.44%, and the stopping criterion is not fulfilled. The robotic agent flags the system to seek assistance, and the system flags back as soon as the obstacle has been removed. When this happens, the robotic agent moves toward the previously blocked area (Figure 6c) and continues exploring the newly generated frontiers until the stopping criterion is fulfilled.

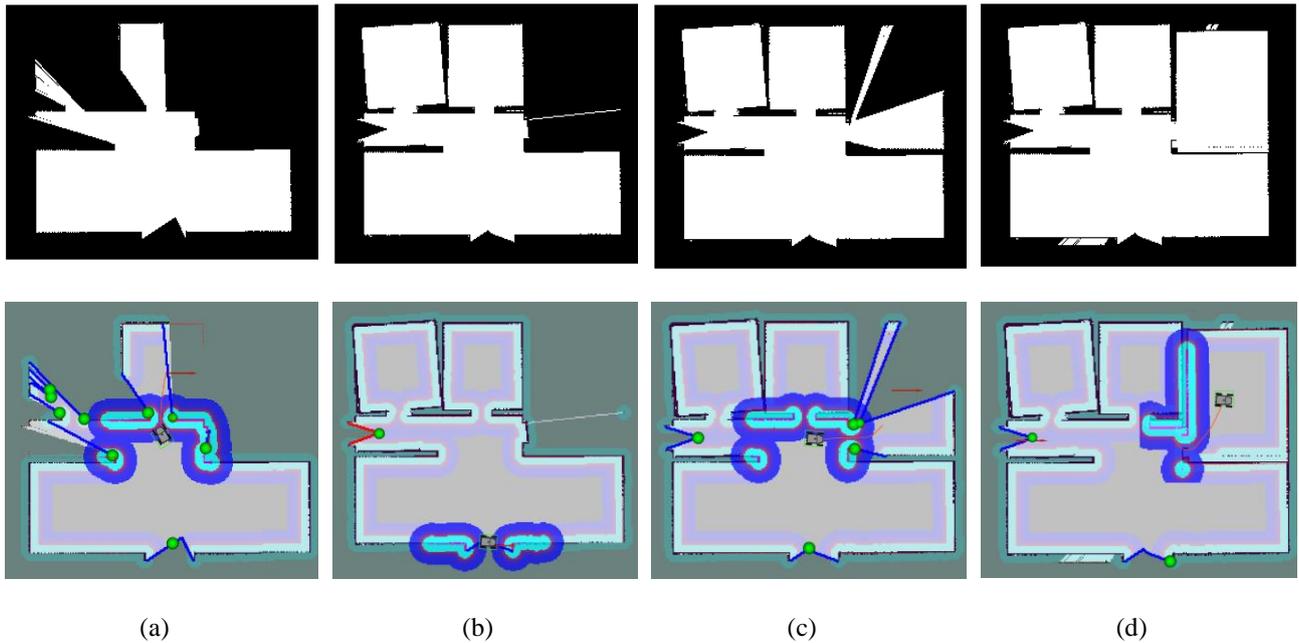

(a)                         (b)                         (c)                         (d)

Figure 6. Binary map of the explored space (upper row) and robot view (lower row) at different stages of the exploration process. (a) stopping criterion of 59.94%, (b) stopping criterion of 79.44%, (c) stopping criterion of 88.11%, (d) stopping criterion of 100.85%.

There are multiple ways in which the different agents that are part of the system can collaborate and help each other as the exploration process continues. The available collaboration methodologies are determined by the morphological features of the agents that comprise the system (i.e., different payloads). Most localization and positioning algorithms are based on sensor input, such as data from the encoders or cameras (i.e., odometry) and LiDARs. To provide accurate results, the environment needs to be rich in features. This means that the localization algorithms will not perform well in a simple and repetitive environment, such as a long hallway without any distinctive features. This allows the autonomous agents to position themselves fully within the environment. For the same reason, if the environment has very similar open spaces (e.g., in the early stages of the construction process), it can lead to errors during the localization (and, therefore, the mapping) process. To solve this, one (or more) of the agents within the system can act as support to provide global positioning within the world reference system. To do this, each agent is equipped with a relative localization system, such as a motion capture (mo-cap) system, cameras or highly reflective and fiducial markers. As long as there is more than one agent within the Field of View (FOV) of the localizer agent, triangulation can be done, and the overall positioning estimation can be improved with this method. As the exploration evolves, it might happen that not all areas are of the same interest to the supervisor. For some specific scenarios, higher-resolution data might be needed, which can be flagged autonomously or by a human operator supervising the process. When that happens, a suitable agent equipped with the right payload (e.g., high-resolution RGB camera, infrared camera, high-resolution 3D scanner) is flagged and commanded to perform the required task in the selected scenario.

When the stopping criterion is finally fulfilled and all the individual tasks have been completed, the process is considered finished. As a result, the system provides all the data gathered by the individual agents during the specific tasks assigned to them, as well as the global map resulting from the exploration algorithm.

## 4 Case study

This case study illustrates the implementation of the methodology presented in Section 3 to use multi-agent robotic systems and exploration algorithms in construction sites. The goal is to show how different robotic systems, depending on their characteristics, functionalities, and limitations, can be used to collaboratively achieve a specific task with minimal human intervention. To that end, the main objective was to test and evaluate the overall performance of the key parts, particularly the integration of multi-agent systems using exploration algorithms to achieve a given task. The task chosen was the collection of point clouds that could be later used for different applications, including the development of 3D models for visualization purposes, quality inspection or progress monitoring. The actual application of the collected data is beyond the scope of this case study, and therefore the processing of said data is excluded from this study. The focus is on (1) the definition, development, and integration of different robotic systems, (2) the use of exploration algorithms to improve the automation of the task and the study of the right stopping criteria for the exploration, and (3) performing a specific task (in this case, collecting data using a 3D laser scanner).

Data collection is a common task in construction projects and a great candidate for automation. The common way in which data is collected using a 3D laser scanner (with static scanners such as FARO Focus S or Leica BLK360) is by having one or two operators choose the scanning locations, manually installing the scanner, and manually initiating the scanning process. New scanning technologies and devices allow real-time registration of the collected point clouds through dynamic scanning (e.g., Hovermap ST from Emesent). This means that the 3D point cloud is being collected as the robot moves instead of having to stop in predefined locations for the scanning. Of course, the type of scanner to be used depends on the purpose for which the data will be used (hence its quality). In general, the quality of the data from "dynamic" scanners is typically good enough for most applications, such as the overall representation of 3D models and getting a general understanding of the as-built conditions of the scanned space. For example, many studies have shown that having a reliable and updated BIM is the first step to increasing efficiency and productivity in the overall construction process [38]. Therefore, a significant amount of research has been dedicated to creating as-built models from the data collected at the site [39], [40]. This process is commonly called scan-to-BIM and is useful for many applications such as progress monitoring and quality assessment.

<u>Description of the experiment</u>

The setting for this case study includes a space of approximately 80 m$^2$ (~860 ft$^2$) with two distinct and separated areas (i.e., separated by clutter and an obstacle). The space comprises semi-finished (exposed concrete and CMU walls) and finished conditions (fully fitted lab and office area) on a university campus. The location was chosen due to the accessibility of the authors to the space and the ability to use the different elements required to test the proposed methodology.

Since one of the objectives is to limit the amount of human intervention during the data collection, high consideration was given to robotic systems (agents) capable of doing their tasks with as much autonomy as possible. With that in mind, the general description and characteristics of the experiment for the data collection process are summarized below:

The ultimate goal is to acquire a complete 3D point cloud data of the entire space. For this particular case, only the area to be explored is known, and there is no prior information about the layout of the space, and no BIM is available. The requirements of the experiment include: (1) generating a ground-truth map representing the layout of the scene, (2) collecting 3D point cloud data of all the areas, and (3) being able to access all areas. All the requirements should be achieved without a human being physically present in the site (i.e., remote interventions might be possible). That means that robotic systems (agents) should be able to handle their tasks autonomously, or at least with very limited intervention from human operators (e.g., remote access or teleoperated intervention could be considered).

With that in mind, to satisfy the first requirement, a ground-truth map was generated by performing SLAM with the robot being manually teleoperated through the environment. It is worth mentioning that this process only needs to happen once or every time the environment significantly changes. Ideally, this process is entirely replaced by having prior information in the form of a BIM. Figure 7 represents the ground-truth map obtained during the SLAM process.

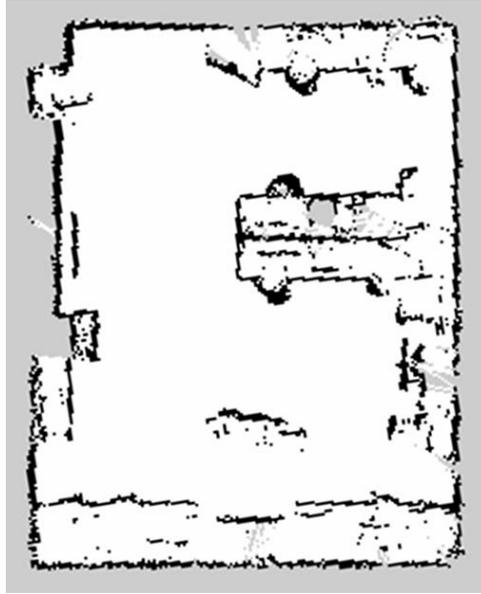

Figure 7. Ground-truth map of the space generated by performing SLAM with the robot teleoperated.

To satisfy the second and third requirements, it is necessary to define the roles of the different agents and select them based on the required sensors, actuators, peripherals, and equipment able to carry out the task (i.e., collect data) in all areas with limited human involvement (and no human presence).

It is assumed that limited information about the space is known (only the total area of the space for which data needs to be collected) and that two robotic systems (agents) will be used. One robotic agent (RA) will be able to generate the navigation map using exploration to reduce the manual input or intervention from humans (instead of manually teleoperating the robot). The exploration algorithm generates this map in real time, ensuring that the robot traverses the entirety of the space, digitizing the 3D model as it navigates the environment. Relying on the exploration algorithm to map the scene instead of trusting the prior information of the space (i.e., ground truth map, floor map) ensures that the robot is aware of the latest state of the scene. Construction environments are very dynamic, and the presence of obstacles would not be reflected on said maps. Due to payload constraints, the robot will not have the capacity for additional equipment. This RA will be able to partially achieve requirements 2 and 3; however, when access to certain areas is not possible (i.e., due to occlusions associated with a typical construction environment), assistance will be required. In that case, a second RA will assist the first RA by opening doors or removing obstacles so that the first RA can continue with its task. Having the assistance of the second RA will allow the full completion of requirements 2 and 3.

The key characteristics and roles of the robotic systems are summarized in Table 1.

Table 1: Roles and main characteristics of the robotic systems used in the case study

| Robot ID | Role | Characteristics |
|---|---|---|
| RA1 | To generate real-time navigation map and collect 3D data using a 3D scanner. | Exploration algorithm, "dynamic" scanner for data collection, communication device, navigation LiDAR. |
| RA2 | To assist RA1 in accessing other areas by interacting with the environment (e.g., opening doors and/or removing obstacles to allow access) | Robotic arm/manipulator, communication, navigation LiDAR. |

## 4.1 Multi-agent robotic system

In this case study, two robotic agents have been chosen to complete the task. Their characteristics and specification are described below.

### 4.1.1 Robotic Agent 1 (RA1)

The Robotic Agent 1 is a four wheels ground robot based on the Robotnik SUMMIT-XL robotic platform, which has been modified and carries the necessary payloads to perform the allocated tasks (Figure 8). The robot is equipped with mecanum wheels that allow it to move omnidirectionally and change the heading (orientation) independently. This functionality gives the ability to the mobile robot to operate in confined spaces where delicate motion maneuvers are required. The RA1 is also equipped with two long-range 3D LiDARs and a short-range depth front-facing camera needed for perception. The main payload of RA1 is a LiDAR-based Hovermap ST 3D scanner from Emesent which can 3D scan dynamically (as it moves) with a range of up to 100 meters. At the end of each scanning session, the scanner provides a full point cloud model with all scanned frames registered and ready for post-processing. The registration of the scanned frames relies on the LiDAR-based odometry of the scanner, which makes it ideal for GPS-denied environments. The second 3D LiDAR is an Ouster OS1 with a maximum range of 150m and a 45º vertical field of view. This Ouster LiDAR is used for navigation purposes, whereas the Emesent LiDAR is used for the case study-specific task (i.e., the 3D digitization of the environment).

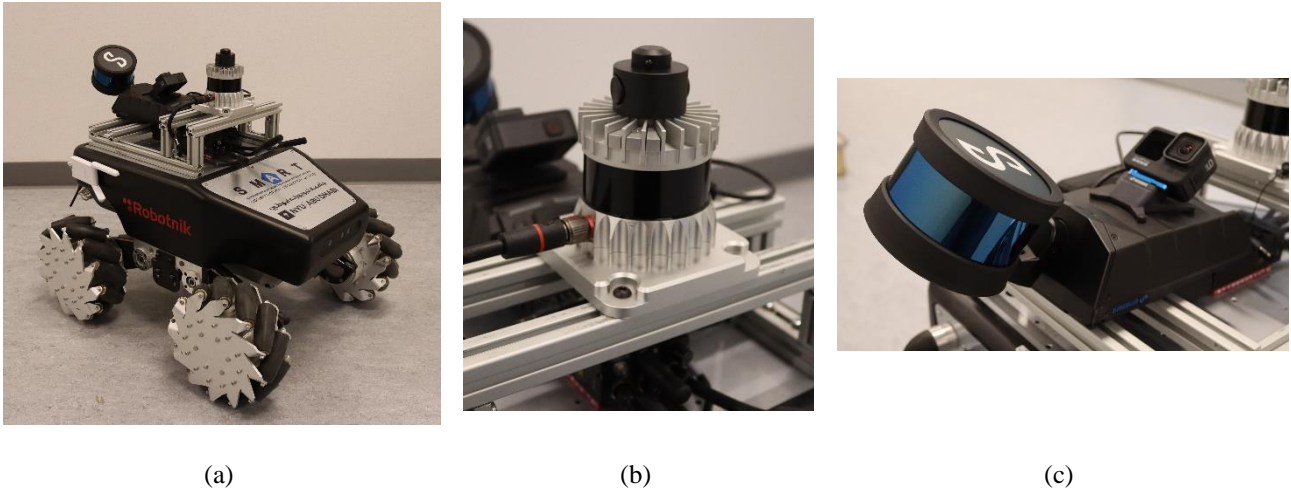

(a)                                        (b)                                        (c)

Figure 8. (a) RA1 with all its payload. (b) OUSTER OS1 LiDAR for navigation. (c) Hovermap ST LiDAR for data collection.

The exploration, navigation, and data acquisition algorithms are running on an x64 architecture on board the Computation Unit inside the robot using Robot Operating System (ROS) as middleware.

The last component of the RA1 is a Communication Module from Persistent System (Figure 9) that creates a Mobile Ad Hoc Networking (MANET) environment to support data rates up to 120 Mbps. Some of the advantages of this Communication Module are that the parameters of the physical communication layer, such as the Transmission Frequency, Transmission Power, Bandwidth, and Multiple-Input Multiple-Output (MIMO) topology, can be configured to adapt to the operating environment. In addition, it supports hardware encryption to secure the transmitted data. For this case study, the parameters of the physical communication layer of the radios were chosen to archive a high-speed stable network. These parameters are summarized in Table 2.

Table 2. Radio transmission parameter for this case study

| Parameter | Value |
| --- | --- |
| Transmission Frequency | 2.420 GHz |
| Transmission Power | 100 mw |
| Bandwidth | 20 MHz |
| MIMO topology | 3X3 |
| Encryption | CTR-AES-256 |

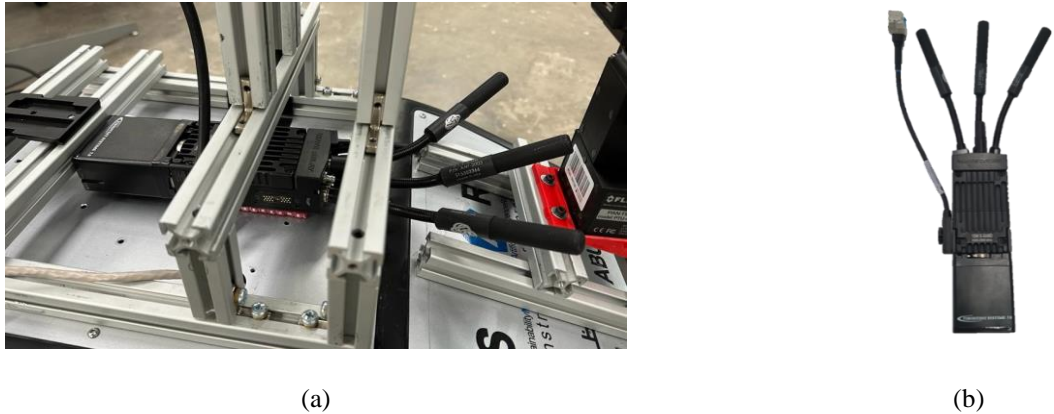

| (a) | (b) |

Figure 9. Communication Module installed on RA1. (a) Location in the RA1 and (b) isolated Communication Module

### 4.1.2 Robotic Agent 2 (RA2)

The RA2 is a four-legged ground robot based on the Boston Dynamics Spot robotic platform, which has been modified to carry the necessary payloads to perform the allocated tasks (Figure 10). The RA2 is capable of navigating in a dynamic construction environment characterized by obstacles, narrow pathways, stairs, etc. The platform is very agile, and it can reach areas that are not possible with the RA1.

The payloads of the RA2 include a high-performance Computing Unit based on an x64 architecture CPU and a CUDA-capable GPU, a six DOF robotic arm with an end effector of a two-finger gripper, and an RGB-zoom camera and thermal camera on a 3-axis gimbal. The robot is also equipped with five short-range depth cameras needed for perception and a communication module, as described in RA1's payloads section.

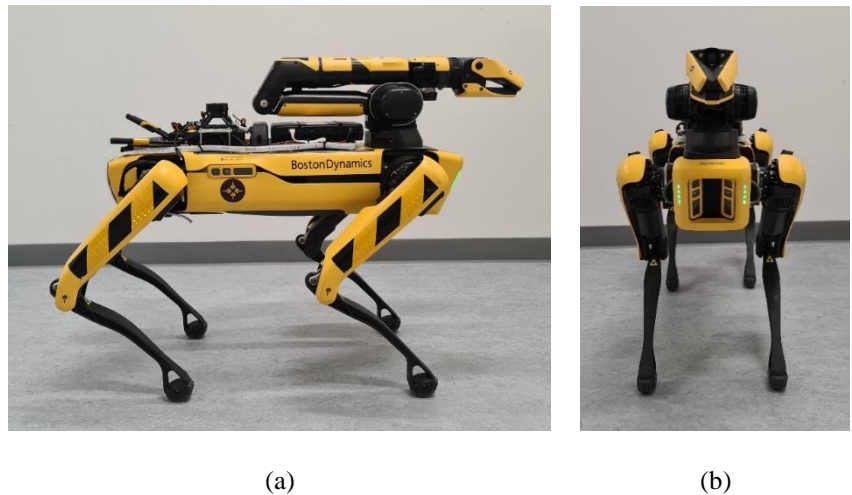

| (a) | (b) |

Figure 10. (a) Side view of the RA2, and (b) front view of the RA2.

### 4.1.3 Control Monitor Station (CMS)

The CMS for this case study consists of two computers responsible for the teleoperation of the Robotic Agents and a Communication Module (CM) as the one described in RA1's payloads section. These computers can be used to allow communication between the RAs with the HAs and do not need any particular specifications since they are only used for communication purposes. This communication works both ways, allowing the HAs to visualize and monitor the progress in real-time and give commands to the RAs, if necessary. Each agent, as well as the different computers in the CMS, are connected to one CM that allows them to be part of the main communication network and communicate with each other. A generic representation of this is illustrated in Figure 11.

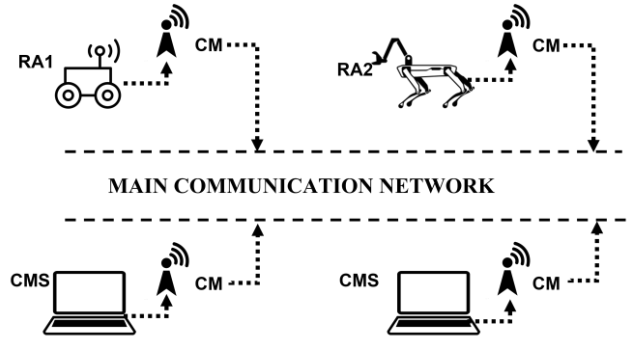

Figure 11. Diagram of communication among different agents.

### 4.1.4 ROS integration

ROS is a structured system comprised of nodes exchanging messages through topics and communicating with each other through services. Each node is in charge of a specific task within the ROS network (e.g., sending velocity commands to the motors, autonomous navigation, processing data from the LiDAR, etc.). Figure 12 shows a simplified ROS network showcasing the main nodes used in this case study. It can be seen how the nodes (ovals in Figure 12) responsible for the data acquisition (i.e., Ouster OS1 LiDAR, Hovermap) share their information in the form of topics (rectangles in Figure 12) (i.e., /pointcloud_data, /3D_scanner_data) with the other parts of the system. Sometimes, the information needs to be processed, for example, converting the 3D information from the OS1 LiDAR to a 2D laser topic (i.e., /2D_laserscan_data) that can be interpreted by the SLAM node. Some nodes have a higher level of commands (i.e., navigation and multi_agent_interaction), allowing them to collect data from multiple sources and command the necessary orders to the system. Finally, there are nodes responsible for commanding the robot to perform specific actions, such as moving the wheels (i.e., robot control). Details regarding the nodes responsible for navigation, localization and SLAM are out of the scope of this study, since they are publicly available within the ROS framework and no new developments have been performed on them.

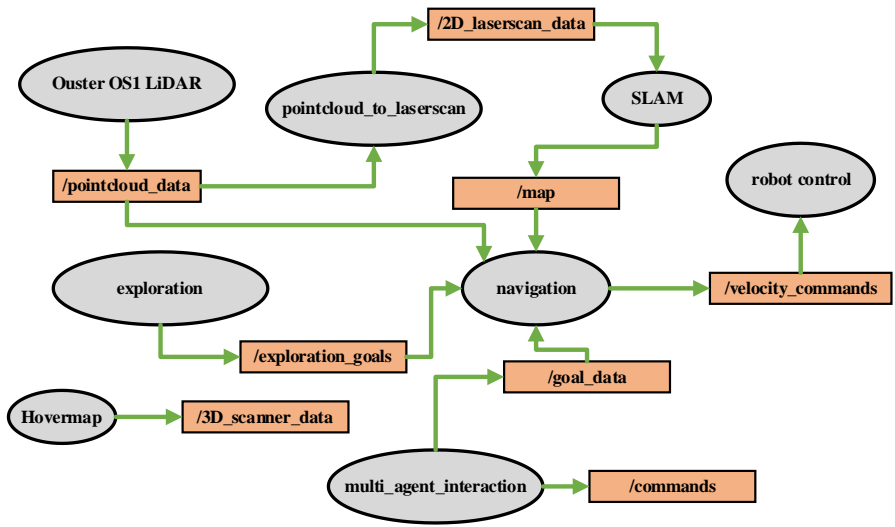

Figure 12. Simplified version of the ROS network used for this case study. All the nodes communicate constantly with the master node.

One of the key aspects of a ROS network is the presence of a master node. The ROS master node is essential to coordinate and supervise all the communication and data exchange in the network. When a node needs to request information from another node, it first contacts the master to ask for the proper routing, and the information gets handed over. This allows the ROS network to be expandable, even beyond the boundaries of a single device. As long as a ROS node is aware of where the master node is and can communicate with it (i.e., within a network), it does not matter where this ROS node is being executed. That means this is a very suitable architecture for a multi-agent approach like the one presented in this study. This architecture allows for any agent (e.g., agent *i*) within the network to communicate (i.e., receive and send information) with any other agent (i.e., agent *n*) in the same network. The need for this master node is also a limitation, in the sense that if this master node was to become unavailable, the entire network would fail and the agents would not be able to function properly. This shortcoming is solved in ROS 2, an upgraded and revised version of the original ROS framework. However, ROS 2 is still in early development stages and is not as stable and reliable as ROS 1.

In particular, the ROS network facilitates the fact that the system can use a series of local maps for each agent to be generated during the exploration that can be shared amongst any other agent and be put together in a single global map, used for positioning all the individual agents within the world coordinate system. This also allows any agent to send an open help requests to the network to exploit the multi-agent feature of the system. For this, a developed node constantly listens to a topic solely dedicated to help requests. A ROS custom message has been developed for this, containing the ID of the agent making the request, its coordinates within the world coordinate system and the nature of the request. Once new information is published on that topic, the node leverages what type of help is being requested and which agent within the network might be suitable to fulfill the request. Once the correct agent has been chosen, a command is sent for this agent to stop the exploration process and attend to the request and the solicited location.

### 4.2 Results/output

As described in the Methodology Section, RA1 begins the exploration approach with a predefined starting pose. The explored area is compared in real time with the ground-truth map obtained beforehand, and the stopping criterion is computed. Figure 13 shows three different moments of the whole exploration process. Figure 13b represents the robot checking all the available frontiers until no feasible waypoint is left, which means the robot has no traversable space left to explore. However, at this point, the percentage of the area explored ($\varphi$) is only 84.53%, which does not satisfy the stopping criteria.

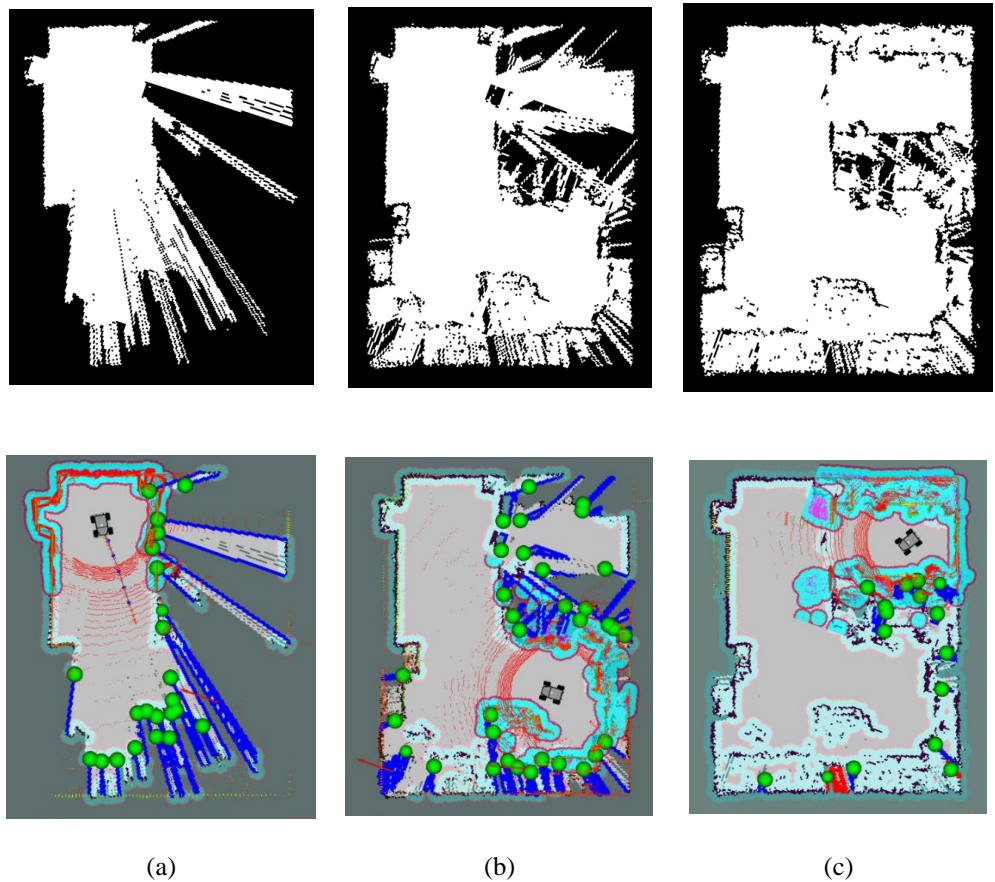

|     |     |     |
| --- | --- | --- |
| (a) | (b) | (c) |

Figure 13. Binary map of the explored space (upper row) and robot view (lower row) at different stages of the exploration process. (a) stopping criterion of 49.54%, (b) stopping criterion of 84.53%, (c) stopping criterion of 95.79%.

Since the stopping criterion has not been fulfilled, RA2 is flagged for help. Since the previous information on the space is present and available to the system, the missing space location is easily obtained and sent to RA2 to check on the blockage. At this point, the RA2 takes a picture with its front camera (Figure 14), and a human agent (HA) remotely indicates a grasping point on the obstacle. Since the nature of the obstacle is hard to predict beforehand, this part of the process is the only one that requires human intervention. After the grasping point is indicated, the RA2 removes the obstacle and clears the path for the RA1 to keep exploring.

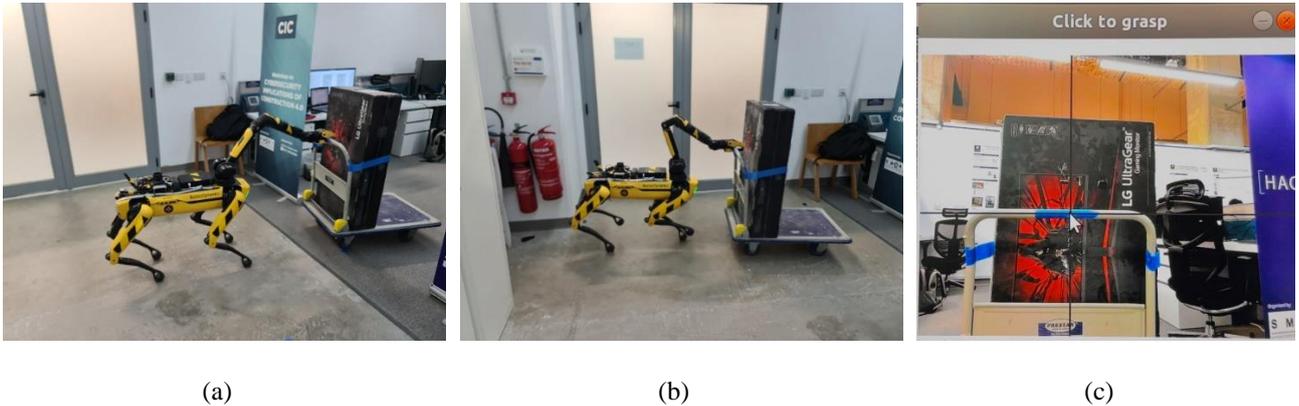

(a)  (b)  (c)

Figure 14. (a) RA2 grasping the obstacle to be removed. (b) RA2 moving the obstacle away from the access. (c) Snapshot of the RA2 camera view.

The map generated by the exploration algorithm still has the blocked path flagged as an obstacle. In order to clear the obstacle from the map, the RA1 is sent to the same location where the RA2 removed the obstacle until the mapping algorithm automatically clears the non-present obstacle from the map (Figure 15). Once the obstacle is cleared, the remaining frontiers to be explored (or the newly generated frontiers) are evaluated, and the RA1 is sent inside the unexplored space. For this experiment, after entering the unknown space, the stopping criterion is satisfied (95.79%) and the exploration process finishes.

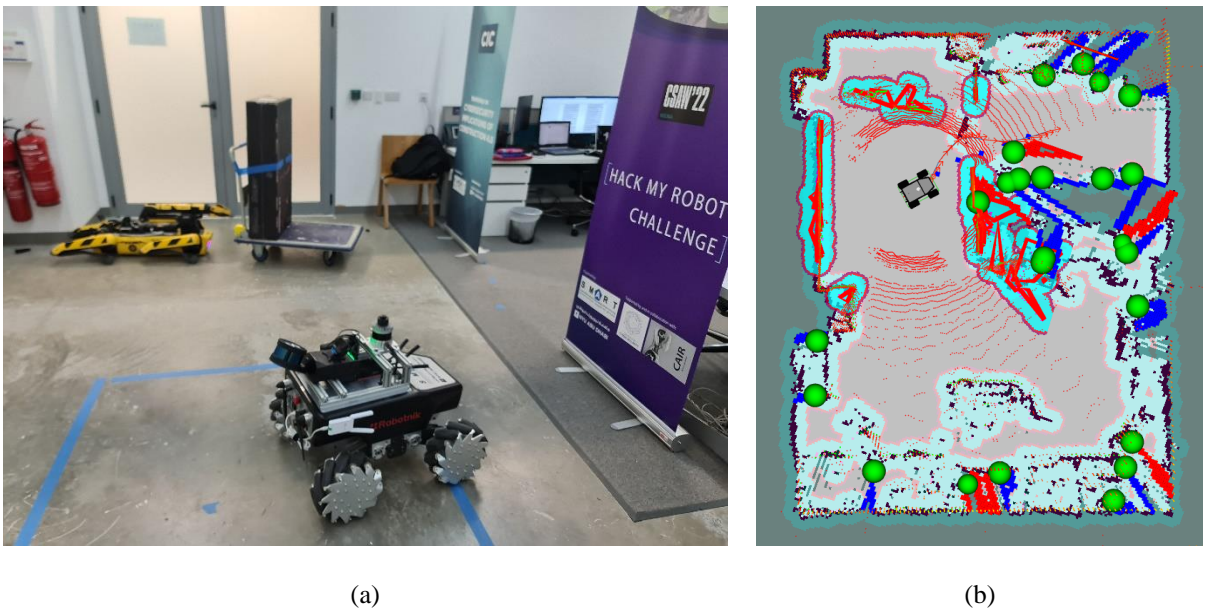

(a)  (b)

Figure 15. (a) World view and (b) robot view of the RA1 planning its way inside the previously blocked zone after RA2 has cleared the path.

The data collected during the exploration process are shown in Figure 16. A video of the experiment described in this section can be found in [41].

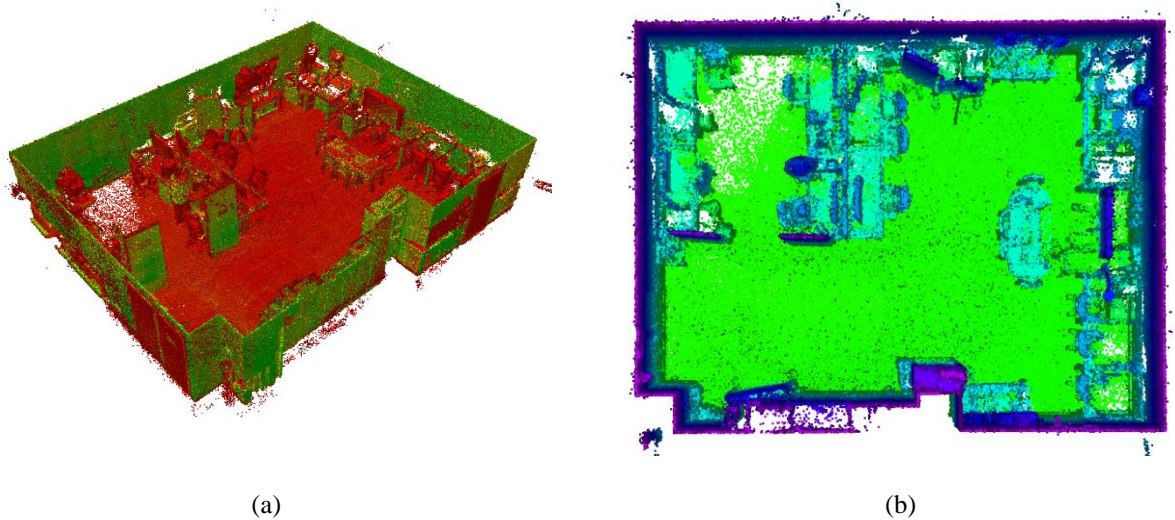

(a)                                                    (b)

Figure 16. (a) Perspective view of the 3D acquired point cloud during the process. The colors represent the normal directions of the points. (b) Top view of the 3D acquired point cloud during the process. The colors represent the elevation of the points, green being the floor.

## 5    Discussion and limitations

As discussed previously, having multiple agents with different capabilities that can support each other (i.e., require collaboration) is needed to achieve an optimal system able to fulfill the given task. Although having human agents makes the process not entirely automated, their presence in the loop is important, especially for non-trivial decisions. Also, part of the process might require some human intervention, either to provide further instructions to the system or to perform a task that none of the robotic agents is suitable to do. Overall, the presence of the human agent makes the system more flexible and adaptable without sacrificing the automation part. In the presented case study specifically, the human intervenes by providing the robot with information regarding an optimal grasping point for removing the trolley. This process could be replaced by applying an Artificial Intelligence algorithm that could easily segment any handles shown in the image. Nonetheless, since the system allows for bidirectional communication, the human agent could still get feedback from the robot if the grasping was unsuccessful and still be present in the loop to help if needed.

Nonetheless, the discussed methodology presents some limitations. For example, the case study used to illustrate the proposed multi-agent robotic system would require to be escalated to fully test the capabilities of the system. Currently, the number of agents is limited (i.e., two robots and one human), and the number of tasks to be performed is reduced (i.e., 3D data collection). A bigger and more complex scenario would be needed to realize the full benefits of the methodology. The presented case allowed to test the principles behind the MARS system, such as the developed communication system between all the agents and the task-allocation procedures. All the system components have been developed with a modular aspect in mind, allowing for full scalability of the project in terms of tasks and agents. Nonetheless, each scenario might present a set of particular challenges that need to be addressed. For example, the communication network stability relies on the distribution of the environment (i.e., big open spaces vs. narrow and long corridors) or its structural composition (i.e., solid concrete walls vs. thinner construction materials). Therefore, the power and frequency modulation would need to be adapted to achieve better results.

As previously discussed, the fact that the ROS network requires a master node to function could be a limitation for the stability of the system. If this master node was to go down for any reason, the entire system would become unavailable.

In addition, prior information from the environment is needed if robust and complete results are to be expected. Prior information can be given in the form of an existing BIM of the building or a previously mapped ground truth of the environment. The presence of prior information could be a challenge for already-built buildings, but new construction projects are expected to have a BIM model in most parts of the world nowadays. Although the proposed methodology could work in a completely unknown scenario, the results might not be as complete as needed for a useful and significant site assessment. Nonetheless, given the modular aspect of the proposed methodology and its scalability, the need for prior information would be reduced as more agents are present in the system.

## 6 Conclusions and future work

The approach presented in this paper aims to solve the lack of automation currently present in the construction field. Due to the complexity and dynamically changing environments characteristic of the construction field, it is especially challenging to develop a single robotic platform that can fulfill the requested task and overcome all the different drawbacks that derive from it. For this challenge, the multi-agent methodology provides a solution by redistributing all the necessary features and characteristics amongst different agents (i.e., robotic or human). Collaboration, either among robots or between robots and humans, will be the key to the successful and robust integration of automated (either semi-automated or fully automated) processes in the construction field.

The methodology has been tested by tasking three different agents (two robots and one human) in a small case study to 3D digitize a small occluded space. Both robotic agents present different payloads and capabilities, and it was proven that collaboration between the different agents was a must to overcome all the challenges that can be present in a complex environment such as a construction site.

Future work to expand the application of the proposed method includes testing in a bigger and more complex environment. Instead of a cluttered lab environment, a real construction site with multiple floors, more agents, and more tasks to be distributed would be used to showcase the full capabilities of the developed methodology. In addition, to reduce the presence of human intervention, different Machine-Learning-based algorithms can be developed to assist with some of the tasks, such as identifying a gripping point to remove an obstacle or the door handle to open a door. Upgrading the entire framework from ROS 1 to ROS 2 would greatly benefit the stability of the system to solve the need of the master node. The authors are currently exploring the possibility of this upgrade, verifying that all the required components are already available in ROS 2.

## Acknowledgments

Part of this research benefited from the resources in the Core Technology Platform (CTP) at New York University Abu Dhabi (NYUAD), particularly the CTP's Kinesis Lab. This work was partially funded by the NYUAD Center for Sand Hazards and Opportunities for Resilience, Energy, and Sustainability (SHORES). This work was partially supported by the NYUAD Center for Interacting Urban Networks (CITIES), funded by Tamkeen under the NYUAD Research Institute Award CG001.